\documentclass[11pt]{article} 
\usepackage{rldm,palatino}
\usepackage{graphicx}
\usepackage{amsmath}
\usepackage{algorithm2e}
\usepackage{hyperref}
\usepackage{caption}
\title{Modeling Attention during Dimensional Shifts \\ with Counterfactual and Delayed Feedback }

\author{
Tailia Malloy \\
Social and Decision Sciences\\
Carnegie Mellon University\\
Pittsburgh, PA 15213 \\
\texttt{tylerjmalloy@cmu.edu} \\
\And
Roderick Seow \\
Social and Decision Sciences\\
Carnegie Mellon University\\
Pittsburgh, PA 15213 \\
\texttt{yseow@andrew.cmu.edu} \\
\And
Cleotilde Gonzalez \\
Social and Decision Sciences\\
Carnegie Mellon University\\
Pittsburgh, PA 15213 \\
\texttt{coty@cmu.edu} \\
}

\begin{document}

\maketitle

\begin{abstract}
Attention can be used to inform choice selection in contextual bandit tasks even when context features have not been previously experienced. One example of this is in dimensional shifts, where new feature values are introduced and the relationship between features and outcomes can either be static (intra-dimensional) or variable (extra-dimensional) \cite{eimas1966effects}. Attentional mechanisms have been extensively studied in contextual bandit tasks where the feedback of choices is provided immediately, but less research has been done on tasks where feedback is delayed or in counterfactual feedback cases. Some methods have successfully modeled human attention with immediate feedback based on reward prediction errors (RPEs) \cite{niv2015reinforcement}, though recent research raises questions of the applicability of RPEs onto more general attentional mechanisms \cite{gershman2024explaining}. Alternative models suggest that information theoretic metrics can be used to model human attention, with broader applications to novel stimuli \cite{malloy2023accounting}. In this paper, we compare two different methods for modeling how humans attend to specific features of decision making tasks, one that is based on calculating an information theoretic metric using a memory of past experiences, and another that is based on iteratively updating attention from reward prediction errors. We compare these models using simulations in a contextual bandit task with both intra-dimensional and extra-dimensional domain shifts, as well as immediate, delayed, and counterfactual feedback. We find that calculating an information theoretic metric over a history of experiences is best able to account for human-like behavior in tasks that shift dimensions and alter feedback presentation. These results indicate that information theoretic metrics of attentional mechanisms may be better suited than RPEs to predict human attention in decision making, though further studies of human behavior are necessary to support these results.
\end{abstract}

\startmain 
\section{Introduction}
How does our experience and the feedback we receive from it impact the way we attend to new situations? Dimensional attention - selectively focusing attention to a relevant subset of dimensions - has been proposed to enable rapid generalization over the space of possible options, allowing humans to efficiently learn and predict the value of both experienced and yet-to-be experienced decisions. However, less is known about how the temporal presentation of feedback impacts our attention mechanisms. This is a critical aspect of human attention in the real world as we often experience long delays before receiving feedback such as in the case of long-term investments. Another example of temporally distinct feedback is in counterfactual cases such as the stock market, where we can observe what would have happened if we had behaved differently. This work seeks to test different possible explanations for human attention in tasks that transfer the decision making dynamics through domain shift, as well as alter the presentation of feedback. 

One possibility is to leverage reward-prediction-error (RPE) to guide attention. RPE reflects the discrepancy between the expected and received outcomes, and has been shown to correlate strongly with the activity of dopaminergic neurons when learning utilities. RPE is also the key quantity that drives learning in the reinforcement learning (RL) framework and has been applied to predict human attention to specific dimension features in contextual bandit tasks \cite{niv2015reinforcement}. However, recently there has been research that suggest that RPEs and iterative updating may not be enough to capture the full range of human attention in utility-based decision making tasks \cite{gershman2024explaining}.

Another possibility is that dimensional attention is determined using a memory of past experiences by tracking the mutual information between dimensions and outcomes. Mutual information reflects the amount of certainty of the value of a variable that is gained by knowledge of another variable. Prior research has found that mutual information is a key driver of category and causal learning \cite{feldman2021mutual}. Thus, we also sought to test whether calculating mutual information between dimensions and outcomes is sufficient in predicting human-like attentional mechanisms. Our proposed method leverages the Instance-Based Learning (IBL) framework, which proposes that option values are computed at decision time by referring to a memory of past experiences \cite{gonzalez2003instance}. 

In this work, we begin by describing the contextual bandit task that varies the type of dimensional shift between intra-dimensional and extra-dimensional, as well as the feedback presentation between immediate, delayed, and counter factual (see Figure \ref{fig:Environments}). We then describe the RL and IBL frameworks as well as their respective methods for predicting attentional weights, referred to as WRL and WIBL. Following this, we present simulations of both RL and IBL models with and without attentional weight learning, and compare their performance with previously studied effects in human decision making. We conclude with a discussion of future research directions for experimentation in human decision making to evaluate the predictions of our proposed IBL weight learning method. 
\section{Dimensional Shift Task}
\begin{figure}[!t]
  \centering
  \includegraphics[width=0.9\linewidth]{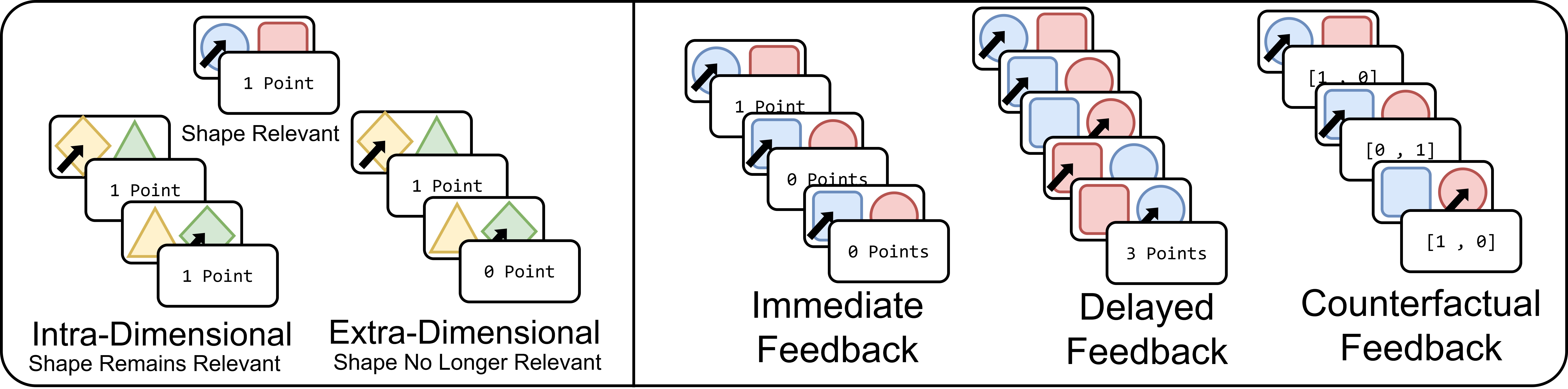}
  \caption{\textbf{Left:} Example of the difference between intra-dimensional where the attribute remains relevant and extra-dimensional shift where the relevant attribute changes. \textbf{Right:} Example of the difference between immediate, delayed, and counterfactual feedback, which vary when and how much choice outcome feedback is presented.}
  \label{fig:Environments}
\end{figure}

The decision making task we use to compare the WRL and WIBL methods is a contextual bandit choice task, which are commonly applied to RL and IBL models \cite{gonzalez2024building, niv2015reinforcement}. This environment consists of three abstract features, each with six possible values. An example of what these abstract feature values can represent is provided in Figure \ref{fig:Environments}, where the two feature dimensions are shape and color. In all trials, selecting the option that contained a specific feature-value pair (e.g red-colored choices) was associated with a higher probability of observing a reward of 1 (75\%), compared to selecting either of the other two choices (25\%). If the choice did not result in the high-valued reward of 1, then the agent would observe a reward of 0. Additionally, to make the decision-making task more challenging, a random Gaussian $\mathcal{N}(0,0.1)$ noise was added to all reward outcomes. 

The first dimension along which the learning environment was varied is whether there was an \textbf{Intra-Dimensional} (ID) shift or an \textbf{Extra-Dimensional} (ED) shift in which a feature value was associated with a high reward in the middle of a training episode \cite{eimas1966effects}. The specific feature value that was associated with a higher probability of observing a reward was consistent for the first 50 trials of a training episode. After this, either the high-reward feature value was changed (e.g red or blue) while keeping the relevant feature (e.g color or shape) static (ID), or changing it (ED). An example of this is shown on the left side of Figure \ref{fig:Environments}, where in the intra-dimensional setting shape is consistently related to reward, and in the extra-dimensional setting the relevant feature shifts from shape to color. 

The second dimension along which the learning environment was varied was the amount and timing of feedback that was provided to the models (shown on the right column of Figure \ref{fig:Environments}). In the \textbf{Immediate Feedback} setting the agent makes a selection of their preferred choice and immediately observed the reward outcome from selecting that option based on the features of the choice they made. In the \textbf{delayed Feedback} setting, the agent observes no immediate reward and instead observes the full sum of 10 choice reward outcomes after each 10 steps of the environment. This setting has less reward information forcing the agent to learn to assign values to their choices based on their observations. Finally, the \textbf{Counterfactual Feedback} setting consists not only of the reward outcome of the selection made by the agent, but also the reward outcomes of the choices that were not selected.
\section{Models}
\subsection{Reinforcement Learning}
RL models learn an optimal policy function $\pi^*(a|s)$ through iterative updating of an approximation of the optimal policy $\pi(a|s) = \max_a Q(a,s)$ using RPEs according to the Temporal Difference (TD) update rule for contextual bandits $Q'(a,s) = Q(a,s) + \alpha (r - Q(a,s))$. This allows the agent to update behavior to maximize future reward observed without keeping an explicit memory of the states visited and the actions performed. In contextual bandit tasks, RL models can speed up the training time learning values associated with individual choice features $V(f)$, rather than the entire state. These choice feature values are updated according to the same TD error update rule, $V'(f) = V(f) - \alpha (r-V(f))$ for each feature in the choice selected. Additionally, RL models can emulate the forgetting properties of human decision making through the addition of a decay parameter $\delta$ that controls the deterioration of values associated with features in choices that are not chosen $V'(f) = \delta V(f)$ for each feature $f$ in the choices that are not selected \cite{niv2015reinforcement}.
\subsubsection{Attentional Weights from Reward Prediction Error}
Attentional weights during domain shifts have previously been modeled using RPEs by keeping a vector of feature weights $\omega_f \in W$ that assigns a relevancy weighting to each feature \cite{dekker2022curriculum}. These weights can then be applied to each feature to adjust the value assigned to different options with the potential for improved learning in cases where a specific feature remains relevant throughout training, giving the weighted feature value function as: 
\begin{equation}
p(c) = \dfrac{\exp(V(c))^\tau}{\sum_{c_i \in S c_i \ne c} \exp(V(c_i)^\tau)} \quad \text{where} \quad V(C) = \sum_{f \in C} w_f V(f) \quad \text{where} \quad w_f' = w_f + \alpha_w (r - w_f)
\label{eq:activation}
\end{equation}
where $\tau$ is the soft-max inverse temperature parameter controlling the probability that the model will select the choice option with the highest value $V(c)$, and $\alpha_w$ is the weight learning speed parameter. These weights allow RL models to speed up post-shift learning when in the intra-dimensional case. We also expect an RL agent using attentional weights to have worse performance in extra-dimensional shifts as the feature is no longer correct, which corresponds to human subject behavior \cite{eimas1966effects}.
\subsection{Instance-Based Learning}
The key feature of IBL models that differentiates them from RL models is the presence of a memory $\mathcal{M}$ that stores instances $i$ composed of choice features $j_i$ utility outcomes $u_i$ and options $k_i$ \cite{gonzalez2003instance}. The values of options are not calculated according to a policy function, but rather are calculated each time an IBL agent observes a new decision-making alternative. These action value predictions $V_k(t)$ are calculated at time $t$ for option $k$ based on a summation of all utility outcomes for instances $u_i$ in memory $\mathcal{M}$ via a weighted soft-max of each instance's activation $A_i(t)$ at time-step $t$. The activation is a cognitive concept \cite{anderson1997act}, calculated by summing a function of a decay term $\delta$ that results in more recent instances being retrieved, a weighted similarity measure $\omega_j (S_{ij} - 1)$ that compares the similarity between instances in memory across each of the $j$ features, and a noise term $\sigma \mathcal{N}(-1,1) = \sigma \xi$ to account for random fluctuations in human memory. This gives the choice value function $V_k(t)$ defined in relation to the activation equation $A_i(t)$ as:  

\begin{equation}
V_k(t) = \sum_{i \in \mathcal{M}_k} \dfrac{\exp{A_i(t)/\tau}}{\sum_{i' \in \mathcal{M}_k}\exp{A_{i'}(t)/\tau}} u_i \quad \text{where} \quad A_i(t) = \ln \Bigg( \sum_{t' \in \mathcal{T}_i(t)}  (t - t')^{-d}\Bigg) + \mu \sum_{j \in \mathcal{F}} \omega_j (S_{ij} - 1) + \sigma \xi
\label{eq:blending}
\end{equation}

The most relevant aspect of these equations for learning attention weights is the method of determining the similarity weights $w_j$. In the past, these weights have typically been set to 1 for all features  \cite{gonzalez2024building}. In this work, we describe a method of learning these weights based on the mutual information between feature values and utility outcomes. 
\subsubsection{Attentional Weights from Mutual Information}
Because IBL models are based on storing instances in memory $\mathcal{M}$ that is iterated at choice time to determine the choice option to select, the weight-learning method can take advantage of the storage and retrieval of past instances to calculate the weights of feature similarities. This results in a prediction of choice that depends on a weighted similarity of past experiences, where the weights of these feature similarities are calculated at choice time based on the agent's experience. 
\begin{equation}
A_i(t) = \ln \Bigg( \sum_{t' \in \mathcal{T}_i(t)}  (t - t')^{-d}\Bigg) + \mu \sum_{j \in \mathcal{F}} \omega_j (S_{ij} - 1) + \sigma \xi \quad \text{where} \quad \omega_j = \dfrac{\exp(I_{\mathcal{M}}(j|u)) ^ {\tau_\omega}}{\sum_{k \ne j} \exp(I_{\mathcal{M}}(k|u)) ^ {\tau_\omega}}
\label{eq:activation}
\end{equation}
where $I_{\mathcal{M}}(j|u)$ is the mutual information of the feature $j$ and the utility outcome $u$, calculated over the entire memory $\mathcal{M}$, and $\tau_\omega$ is the mutual information weight soft-max temperature parameter. This method of calculating feature weights is distinct from the RL method of learning weights based on RPEs, as it is not iteratively updated and kept as a separate component of the model. While it is possible to use an iterative update technique to approximate mutual information in RL for contextual bandit tasks \cite{malloy2023learning}, we focus on using the entire memory for IBL models since it is available to the agent. 
\subsection{Simulation Results}
\begin{figure}[!t]
  \centering
  \includegraphics[width=\linewidth]{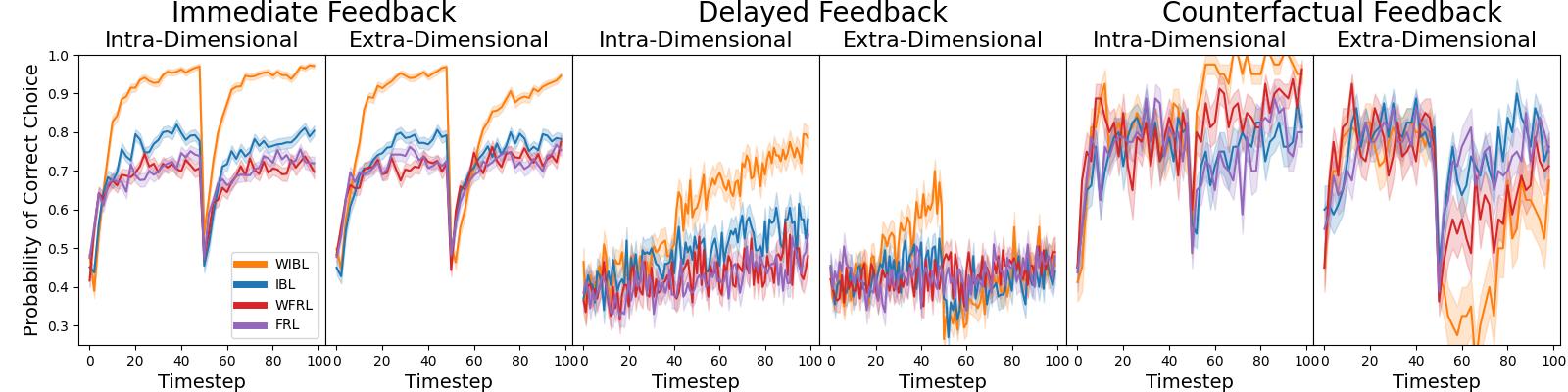}
  \caption{For all plots the shaded region represents one standard deviation. The probability of the model selecting the correct (higher expected utility) option is compared over 100 time-steps with the domain shift occurring at the 50th time-step. For each of the left, middle, and right main columns the left plot is the intra-dimensional domain shift setting, and the right is the extra-dimensional shift setting. The left main column is the immediate feedback setting, the middle is the delayed feedback setting, and the right is the counterfactual feedback setting.}
  \label{fig:Results}
\end{figure}
To gather experimental simulation results, IBL and RL models with and without the weight learning methods described above were compared across the three types of feedback (Immediate, Delayed, and Counterfactual) and the two types of dimensional shift (Intra-Dimensional and Extra-Dimensional). All training episodes consisted of 100 training time steps with the dimensional shift occurring at the 50th time step. 200 agents were trained for each type of model in each pair of feedback and shift types, for a total of 6400 simulated agents. IBL and RL models, simulations, 
and plotting code is available online\footnote{\url{https://github.com/TylerJamesMalloy/dimensionalAttention}}. For IBL models, all parameters (decay $\delta$, temperature $\tau$, noise $\sigma$, and mismatch penalty $\mu$) are set to predefined default values determined by a variety of IBL model experiments \cite{gonzalez2024building}. For RL models, parameters are set to mean values from a previous experiment on a contextual bandit task without a domain shift but with the same number of features and values, as well as the same reward structure \cite{niv2015reinforcement}.

The first comparison of simulation results shown in Figure 2 is in the immediate feedback version of the dimensional shift task between the FRL and IBL methods with and without their respective weight-learning methods. In the intra-dimensional shift, the feature dimension remains the same after the shift, compared to the extra-dimensional shift in which the dimension changes. Thus, we would expect a human learning in this task to have a decreased performance in the first few trials (Jumpstart Performance) after a dimensional shift in the extra-dimensional setting, as confirmed by studies of human subject decision making in this type of task \cite{eimas1966effects}. However, only the WIBL model displays this difference in Jumpstart Performance. Interestingly, while the WIBL model displays worse performance in the extra-dimensional shift initially, it dominates all other model's performance at the end of training. 

The second comparison of simulation results shown in the middle column of Figure \ref{fig:Results} is the delayed feedback setting, in which the feedback is only provided to the agent after 10 selections in the task. The expected result found in human subjects is a slowed learning compared to immediate feedback. To the best of our knowledge there has been no human subject experiment in this type of task that combines delayed feedback with a dimensional shift. These results demonstrate that the WIBL model has higher performance in the intra-dimensional shift, and a more significant deterioration after the shift in the extra-dimensional shift, though the asymptotic performance of all models is comparable. 

The final comparison we make in Figure \ref{fig:Results} is on the counterfactual feedback results shown in the right column of Figure \ref{fig:Results}. Here, the most obvious difference in learning curves is that in the extra-dimensional task the IBL model with weight learning has a significant decrease in performance after the domain shift, compared to the other models which have less of a decrease. Additionally, the IBL model with weight learning shows a faster increase in performance in the intra-dimensional task compared to the other models. This makes intuitive sense, as the counterfactual feedback allows the IBL model to calculate the mutual information of each weight feature as if it has had more experience in the environment, meaning the feature weights are closer to their true values after even a short period of learning.
\section{Future Work}
The results in this work demonstrate that the our proposed WIBL method of updating IBL model weight parameters using a mutual information metric most closely replicates known effects in human decision making in contextual bandit tasks. Additionally, these simulations provide predictions for how humans would behave in two as of yet unstudied decision making settings. Firstly, these simulations predict that in delayed feedback contextual bandit settings with a dimensional shift, there is an exaggerated effect of extra-dimensional shifts compared to intra-dimensional shifts on jumpstart performance. Secondly, the counterfactual feedback setting in this same environment also shows an increased effect of the difference between extra-dimensional shifts and intra-dimensional shifts on jumpstart performance. However, in the delayed feedback setting there is no decrease in performance immediately after the shift, whereas for counterfactual there is. Whether or not weight learning eventually leads to improved asymptotic performance depends on the timescales of the episodes. We plan future experiments to test these predictions in domain shifts with altered feedback.

\textbf{Acknowledgements:} This research was sponsored by the Army Research Office and accomplished under Australia-US MURI Grant Number W911NF-20-S-000, and the AI Research Institutes Program funded by the National Science Foundation under AI Institute for Societal Decision Making (AI-SDM), Award No. 2229881.

\bibliographystyle{unsrt} 
\bibliography{rldm}

\end{document}